\documentclass[conference]{IEEEtran}
\IEEEoverridecommandlockouts

\newif\ifExtendedVersion
\ExtendedVersiontrue

\usepackage{cite}
\usepackage{textcomp}
\usepackage{microtype}

\usepackage{graphicx}
\usepackage[table]{xcolor}

\usepackage{booktabs}
\usepackage{multicol,multirow}

\usepackage{algorithm}
\usepackage{algpseudocode}

\usepackage{amsmath,amssymb,amsfonts,amsthm}
\usepackage{mathtools}

\usepackage{pgfplots}
\pgfplotsset{compat=1.12}
\usepackage{adjustbox}
\newcommand{\ms}[2]{#1\,$\pm$\,#2}

\usepackage[textsize=tiny, textwidth=0.55in, colorinlistoftodos, disable]{todonotes}

\usepackage[hidelinks, breaklinks]{hyperref}
\usepackage[capitalize, noabbrev]{cleveref}

\begin{document}

\setlength{\marginparwidth}{0.53in}

\title{SAGE: Similarity-Based Cleaning of\\Poisoned Training Data from Verified Examples}

\author{
\IEEEauthorblockN{
Chaeeun Han\IEEEauthorrefmark{1}, 
Soodeh Atefi\IEEEauthorrefmark{2}, 
Yevgeniy Vorobeychik\IEEEauthorrefmark{3}, and 
Aron Laszka\IEEEauthorrefmark{1}
}

\IEEEauthorblockA{\IEEEauthorrefmark{1}\textit{College of Information Sciences and Technology}, \textit{Pennsylvania State University}, University Park, PA, USA}

\IEEEauthorblockA{\IEEEauthorrefmark{2}\textit{Department of Computer Science and Engineering}, \textit{University of Louisville}, Louisville, KY, USA}

\IEEEauthorblockA{\IEEEauthorrefmark{3}\textit{Department of Computer Science and Engineering}, \textit{Washington University in St. Louis}, St. Louis, MO, USA}

\IEEEauthorblockA{Email: \{cfh5554, aql5923\}@psu.edu, soodeh.atefi@louisville.edu, yvorobeychik@wustl.edu}
}

\ifExtendedVersion
\IEEEaftertitletext{\centering Published in the Proceedings of the 6th ICDM Machine Learning for Cybersecurity Workshop (ICDM MLC 2026).}
\fi

\maketitle

\begin{abstract}
As machine learning increasingly relies on public, untrusted data sources, data poisoning attacks, which inject malicious examples into training data to induce misclassification of a chosen target, pose a growing threat. Existing defenses either assume zero ground-truth information about which examples are poisoned, or they assume access to a large set of examples verified to be clean. Satisfying the latter assumption incurs significant cost since reliable verification can be very resource- or labor-intensive. 
This cost is particularly high for clean-label attacks, where poisoned examples are visually indistinguishable from clean data. 
Since requiring a large set of verified examples is impractical, we propose relying on a \textit{small set of verified examples including both clean and poisoned ones}, i.e., each example verified either to be clean \emph{or} poisoned through inspection by a forensic expert. 
The challenge is then to detect poisons based on a set of verified examples that is so small that most classification models would  overfit.
To address this challenge, we propose \textit{Similarity-based Approach for Ground-truth-driven Exclusion} (SAGE), which trains a generic feature extractor on a separate dataset and then flags poisoned training examples using a non-parametric, similarity-weighted prediction based on the verified set. %
On standard benchmarks against seven clean-label attack methods, we demonstrate that having access to even a handful of verified poisoned examples provides a substantial advantage. 
We also find that the distribution of verified clean examples across classes matters more than the number of verified examples.
\end{abstract}


\section{Introduction}
\label{sec:intro}

Machine learning (ML) systems have become indispensable in a wide range of applications, from autonomous driving~\cite{jiang2020poisoning} to health care diagnostics~\cite{javaid2022significance}. The success of these systems hinges on the integrity and quality of their training data. However, as ML increasingly relies on external, untrusted data sources, it becomes more vulnerable to data poisoning attacks, in which adversaries introduce malicious examples into training data to cause misclassifications of specific test instances~\cite{tian2022comprehensive,wang2022threats,cina2023wild,fan2022survey}.
Such attacks can be practical:
Carlini et al.~\cite{carlini2024poisoning} demonstrate that an adversary could have succeeded in achieving their misclassification goal by controlling a mere 0.01\% of a web-scale dataset such as LAION-400M.

A number of cleaning approaches have been introduced to detect and remove poisoned examples from training data~\cite{cina2023wild, tian2022comprehensive},
which vary in terms of their assumptions.
Some cleaning approaches trust the class labels of the entire training set~\cite{peri2020deep}; others assume nothing and rely on statistical outlier criteria, at the cost of lower precision~\cite{tran2018spectral, yang2022not}. We focus on \textit{base-set defenses} that assume access to a set of verified clean examples, which serve as a reference for the remaining data. 
The assumed size of this verified set varies between cleaning approaches: the frequency-based poison detector~\cite{zeng2021rethinking} and MW-Net~\cite{shu2019meta} use 100 clean examples; while MNTD~\cite{xu2021detecting}, Neural Cleanse~\cite{wang2019neural} and I-BAU~\cite{zeng2021adversarial} use 1,000. De-Pois~\cite{chen2021pois} assumes 20\% of the training data to be verified, which corresponds to 10{,}000 verified images for CIFAR-10, a commonly used benchmark. 
Reducing the verified fraction degrades both its accuracy and the F1 score from roughly 0.9 to 0.7 as the fraction decreases to 0.5\%. DUTI~\cite{zhang2018training} likewise relies on a substantial number of verified clean examples.

However, verifying that an example is clean is expensive. Clean-label poisons are perturbed to remain visually indistinguishable from benign data, and professional inspectors struggle to identify over 86\% of feature-perturbing poisons~\cite{zeng2023meta}. Verification therefore cannot rely on rapid visual screening; it requires deliberate per-example scrutiny, whose cost grows linearly with the size of the verified set. For this reason, gathering the hundreds or thousands of verified clean examples that existing base-set defenses require is often impractical.

Further, the assumption that verification provides only clean examples may also be impractical.
An inspection process that is meticulous enough to reliably verify examples as clean may also confirm some examples to be poisoned, and these carry more information than what clean examples carry alone. A verified clean set characterizes only how clean data are distributed, so a defense built on it must treat every deviation from this distribution as suspect. A set of verified poisons instead marks regions of the feature space that the attack targets, enabling defense to focus on these specific regions. We therefore study a setting in which the defender has access to both kinds of verified examples, and we show that relying on both enables effective detection based on verified sets that are much smaller than those assumed by existing base-set defenses.

Given both kinds of verified examples, the detection of remaining poisons can itself be posed as an ML problem: a binary classification task separating poisoned examples from clean examples, trained on the set of verified examples and applied to the rest of the training data.
For the kind of unstructured data that we consider throughout this work, such as images, training a classifier based on a few dozen verified examples is very challenging. Training a model directly on so few examples immediately overfits the model. Fitting a classifier on features output by a fixed, pretrained feature extractor only shifts the difficulty: clean-label poisons differ from benign data by small perturbations, so any representation exposing them must retain fine-grained detail and is therefore high-dimensional, leaving the decision boundary under-determined, creating a curse of dimensionality for the detection problem. Both failures share a cause: each fits parameters to the verified set, which is far too small to determine them. This motivates an approach that fits no parameters to that set at all.

In this work, we propose \textbf{SAGE} (\textbf{S}imilarity-based \textbf{A}pproach for \textbf{G}round-truth-driven \textbf{E}xclusion), which addresses these challenges. SAGE trains a feature extractor on a separate dataset, then classifies each candidate as poisoned or clean by a non-parametric, similarity-weighted prediction over the verified set. Because no parameters are fit to that set, its size affects only how finely the prediction can discriminate, not the risk of overfitting.
This shifts the technical question from \emph{how to classify} to \emph{how to represent}. An extractor trained with a standard classification objective is optimized to make classes linearly separable, not to place similar examples near each other, which is what SAGE relies on at detection time. We therefore train it with an objective that mirrors detection. Each batch is split into a reference subset and a query subset, predictions for the query subset are formed as a similarity-weighted combination of reference labels, and the resulting cross-entropy is back propagated through the extractor. Detection then applies the same rule with the verified examples as the reference set. %

We evaluate SAGE on CIFAR-10~\cite{krizhevsky2009learning} and Tiny ImageNet~\cite{Le2015TinyIV} against three clean-label triggerless attacks (feature collision~\cite{shafahi2018poison}, bullseye polytope~\cite{aghakhani2021bullseye}, and gradient matching~\cite{geiping2020witches}) and against four clean-label backdoor attacks (Narcissus~\cite{ zeng2023narcissus} and three variants of Wicked Oddities~\cite{nguyen2025wicked}). On CIFAR-10, SAGE finds at least 480 of the 500 poisoned examples in every triggerless setting, limiting attack success rate (ASR) to at most 5\%, while discarding less than 2\% of the training data. Against backdoor attacks, SAGE is the only defense that effectively suppresses attacks without discarding too much training data: the only baseline method that attains significantly lower ASR does so by discarding so much training data that the accuracy of the trained application model falls below 0.50.

The remainder of this paper is organized as follows.
Section~\ref{sec:related_works} reviews related work on data poisoning and defenses, and explains how our assumptions differ.
Section~\ref{sec:threat_model} introduces the threat model and our objective.
Section~\ref{sec:proposed_approach} presents our proposed method, SAGE.
Section~\ref{sec:experimental_setup} describes our experimental setup, and
Section~\ref{sec:results} reports results.
Section~\ref{sec:conclusion} discusses the limitations of our approach and provides concluding remarks.

\section{Related Work}
\label{sec:related_works}

\subsection{Data Poisoning Attacks}

Data poisoning attacks compromise machine learning (ML) systems by injecting malicious examples into training datasets, disrupting model performance~\cite{fan2022survey, goldblum2022dataset, tian2022comprehensive}. Unlike evasion attacks, which manipulate test-phase inputs without corrupting the model itself, poisoning attacks occur during the training phase~\cite{biggio2013evasion}. This fundamental difference allows poisoning attacks to have long-lasting effects on the model's behavior.

Attacks are commonly categorized along two axes. The first distinguishes untargeted attacks, which degrade overall performance~\cite{fan2022survey}, from targeted attacks, which cause specific test instances to be misclassified~\cite{guo2020practical}. 
The second distinguishes label-flipping attacks, which alter labels~\cite{xiao2012adversarial}, from clean-label attacks, which perturb features while leaving the original labels intact~\cite{shafahi2018poison}. Clean-label attacks are the stealthier of the two because each poisoned example remains consistent with its label. The attack survives inspections that would immediately expose a flipped label.

Their practicality has been established at scale. Carlini et al.~\cite{carlini2024poisoning} show that an adversary could have controlled 0.01\% of a web-scale dataset such as LAION-400M or COYO-700M by purchasing expired domains that the dataset indexes, thereby enabling data poisoning attacks.
We consider three representative clean-label targeted attacks: feature collision~\cite{shafahi2018poison}, which optimizes poisons to collide with a target in feature space; bullseye polytope~\cite{aghakhani2021bullseye}, which surrounds the target with a convex hull of poisons; and gradient matching~\cite{geiping2020witches}, which aligns poison gradients with those of the adversarial objective.

\subsection{Defense Strategies against Data Poisoning}

Defenses that filter suspected poisons from the training set~\cite{steinhardt2017certified, koh2022stronger} differ mainly in what they assume the defender already knows. We organize them accordingly.

\paragraph{Label-dependent defenses}
These trust the class labels of the entire training set. Deep $k$-NN~\cite{peri2020deep} flags a point as poisoned when its own label disagrees with the plurality label among its $k$ nearest neighbors in feature space. The assumption suits clean-label attacks, where labels are correct by construction, and Deep $k$-NN is a strong baseline in exactly that regime; it provides no advantages when labels themselves are manipulated.

\paragraph{Ground-truth-agnostic defenses}
These assume nothing, identifying poisons using statistical, outlier-based, or clustering criteria: spectral signatures~\cite{tran2018spectral}, activation clustering~\cite{chen2018detecting}, and gradient-based core-set selection~\cite{yang2022not}. Freedom from assumptions is costly. Meta-Sift~\cite{zeng2023meta} evaluates on five criteria on CIFAR-10 and finds that none of them isolates clean data with sufficient precision, with several performing worse than random selection under at least one attack. In practice, these methods tend toward over-filtering, discarding large volumes of clean data to catch the poisons~\cite{qi2023towards}.

\paragraph{Base-set defenses}
These assume access to a set of verified clean examples, against which the remaining data is compared. Assumed sizes vary widely: the frequency-based poison detector~\cite{zeng2021rethinking} and MW-Net~\cite{shu2019meta} use 100 examples, MNTD~\cite{xu2021detecting}, Neural Cleanse~\cite{wang2019neural} and I-BAU~\cite{zeng2021adversarial} use 1{,}000, and De-Pois~\cite{chen2021pois} adopts 20\% of the training data by default---10{,}000 images on CIFAR-10---reporting that accuracy and F1 fall from roughly 0.9 to 0.7 as this fraction shrinks toward 0.5\%. DUTI~\cite{zhang2018training} likewise relies on a set of trusted examples.

The difficulty with this third group is not the size of the assumed set, but its assumed purity. Zeng et al.~\cite{zeng2023meta} show that across five representative base-set defenses, performance collapses once less than 1\% of the base set is poisoned. For the frequency-based detector, a single poisoned image among 100 drives performance below the random baseline. The verification step that would prevent this is also unreliable for precisely the attacks we study. Across 16 attacks inspected by 48 professional annotators, human inspectors missed over 86\% of poisons that perturb features while leaving labels intact, close to chance. A large presumed-clean set is thus not simply expensive to obtain. At the sizes these defenses assume, contamination is likely, and contamination is what breaks these defenses.

SAGE differs from all three. It does not trust the labels of the untrusted data, does not operate without ground truth, and does not assume a pure base set. Instead, it assumes a small verified set in which examples are labeled as clean \emph{or} poisoned, and uses both.
Relative to label-dependent defenses, this trades a large volume of assumed-correct class labels for a much smaller number of integrity judgments, which survive label manipulation. 
Relative to ground-truth-agnostic defenses, it supplies the anchor those methods lack, at the cost of an assumption they avoid. 
Relative to base-set defenses, it removes the purity requirement, since verified poisons are labeled as such rather than presumed absent. Verification of this kind is also realistic. Data forensics, incident response, and expert or crowdsourced inspection~\cite{tahmasebian2020crowdsourcing} all produce examples of both kinds, yet existing defenses retain only half of what they yield.

\section{Problem Setting and Threat Model}
\label{sec:threat_model}

\paragraph{Setting}
A defender trains a classifier $f_\theta$ on a dataset $D = \{(x_i, y_i)\}_{i=1}^{n}$ with $y_i \in \{1, \ldots, K\}$. 
An attacker corrupts a small subset by replacing $m \ll n$ \emph{base} examples $x_{b_j}$ with \emph{poison} examples $x_{p_j}$.
Depending on the attack, the poison is either an additive perturbation $x_{p_j} = x_{b_j} + \delta_j$ with $\|\delta_j\|_\infty \le \epsilon$, or a localized patch overwriting a small region of $x_{b_j}$. In both cases the modification is small enough that the poison is not obviously anomalous to a human inspector. Labels are left unchanged, $y_{p_j} = y_{b_j}$.

\paragraph{Attacker}
The attacker aims for a model trained on $D'$ to misclassify a chosen set of test inputs $T$ into an attacker-chosen label $y_b$, while behavior on all other inputs and hence overall test accuracy remains largely unaffected. In \emph{targeted} attacks, $T = \{x_t\}$ is a single test example with a true label $y_t \neq y_b$, and no modification is made at test time. In \emph{backdoor} attacks, $T$ is the set of inputs carrying the attacker's trigger patch, so the attacker controls a whole region of input space rather than a single point.

Because labels are not altered, no label inconsistency is obviously visible for a human to catch. The poisoned model appears normal under aggregate accuracy monitoring in both cases: in the targeted setting because only one input is affected, and in the backdoor setting because the trigger is absent from clean test data.

\paragraph{Defender}
The defender observes only $D'$, and knows neither $m$, nor which examples were poisoned, nor $x_t$, $y_b$, or~$\epsilon$, nor which attack was used. The same procedure is applied regardless of the underlying attack.

Beyond the entire possibly poisoned dataset $D'$, the defender has access to a small \emph{ground-truth set} $G = G_{\text{clean}} \cup G_{\text{poison}} \subset D'$ of examples independently verified as clean or poisoned, with $|G| \ll n$. We use $g_j \in \{0,1\}$ for the verification label of example $j \in G$, with $g_j = 1$ indicating a poison, and $q = |G_{\text{poison}}|$ for the number of verified poisons. We assume $G_{\text{clean}}$ is class-balanced, containing $c$ verified clean samples from each of the $K$ classes.

\paragraph{Goal}
The defender produces a cleaned dataset $\hat{D} \subseteq D'$, obtaining parameters $\theta(\hat{D})$. Given an unknown set of poisoned samples $P \subset D'$, the defense should satisfy three requirements at the same time. First, the attack should fail on the attacker's targets, so that $f_{\theta(\hat{D})}(x_t) \ne y_b$. Second, accuracy on cleaned data should be close to what training on the original clean $D$ would have achieved. Lastly, the number of examples discarded, $|D' \setminus \hat{D}|$, should be small. None of $x_t, y_b$, or $P$ is observed.

These three requirements are in tension. Discarding most of~$D'$ can satisfy the first one but fails the others, while discarding nothing may satisfy the last two but fails the first one. The quantity that distinguishes defenses is therefore not attack success alone, but also the amount of data discarded to achieve it, and we report both throughout Section~\ref{sec:results}.

\section{Proposed Approach: SAGE}
\label{sec:proposed_approach}

SAGE (Similarity-based Approach for Ground-truth-driven Exclusion) identifies poisoned training examples by comparing them, in a learned feature space, to a small set of verified poisoned and clean examples. It proceeds in two stages. A feature extractor $\phi$ is first trained on CIFAR-100 so that images sharing a label are placed close together (Section~\ref{sec:training_phi}). At defense time, the frozen $\phi$ is reused to interpolate the verified set's poison/clean designations over that same similarity structure (Section~\ref{sec:defense_application}).

\subsection{Feature Extractor}

Rather than adopting a pretrained vision model, we train a feature extractor from scratch on a separate dataset. The verified set is far too small to train on directly, so the representation must come from elsewhere; and a pretrained model is optimized for a linear classifier on top of its features, whereas SAGE compares examples by cosine similarity. 
Training our own extractor also lets us set the output dimensionality directly, which we tune for similarity-based detection 
(\ifExtendedVersion%
Appendix~\ref{sec:hyperparameter_search}%
\else%
Appendix A in the extended online version~\cite{han2026sage}%
\fi).

The extractor maps an image $\mathbf{x}$ to a feature vector $\mathbf{z} = \phi(\mathbf{x}) \in \mathbb{R}^{d}$ by adding a two-layer head to a ResNet-18 backbone:
\begin{equation}
\label{eq:phi}
\mathbf{z} = \phi(\mathbf{x}) =
  \mathbf{W}_2\, \sigma\!\big(\mathrm{BN}(\mathbf{W}_1\, \mathbf{r}(\mathbf{x})
  + \mathbf{b}_1)\big) + \mathbf{b}_2,
\end{equation}
where $\mathbf{r}(\mathbf{x}) \in \mathbb{R}^{512}$ is the ResNet-18's pooled feature, $\sigma$ is a sigmoid, and $\mathrm{BN}$ is batch normalization. The head maps $\mathbf{r}(\mathbf{x})$ to a hidden width $h$ and then projects to the output dimension~$d$, the space in which we calculate similarity.
Alongside~$\phi$, we also learn a scale parameter $\tau > 0$, which controls how sharply the prediction (Equation~\eqref{eq:prediction}) concentrates on the most similar references. 

\subsection{Training the Feature Extractor}
\label{sec:training_phi}

Each training batch is split into a \emph{reference} set $R$, whose labels are treated as known, and a \emph{query} set $Q$, whose labels must be predicted from $R$ alone by similarity. A query sample is classified correctly only when same-label reference samples receive high similarity weight. Since that weight comes from distance in feature space, minimizing the loss forces $\phi$ to place same-label samples close together.
At detection time the same extractor and the same prediction rule are reused, with the verified set in place of $R$ and verification labels in place of class labels.

The extractor is trained on CIFAR-100~\cite{krizhevsky2009learning}, disjoint from CIFAR-10 and Tiny ImageNet, the datasets attacked in our experiments (Section~\ref{sec:experimental_setup}); this separation ensures the defense is not biased by the datasets under attack. We expect the resulting space to transfer reasonably to both targets: all three are standard image benchmarks with broadly similar visual statistics, and training against 100 diverse classes rather than a handful discourages $\phi$ from specializing to a few categories---the same reasoning behind why models pretrained on large, diverse label sets transfer to unrelated downstream tasks.

\paragraph{Training Step}
Each training step consists of three sub-steps.
\textit{1. Similarity calculation:}
Every image in $R \cup Q$ passes through $\phi$. For a query image $i$ and a reference image~$j$, we compute the cosine similarity between their features,
\begin{equation}
\label{eq:sim}
\mathrm{sim}(\mathbf{z}_i, \mathbf{z}_j)
  = \frac{\mathbf{z}_i^\top \mathbf{z}_j}
         {\|\mathbf{z}_i\|\,\|\mathbf{z}_j\|}.
\end{equation}

\textit{2. Prediction:}
The predicted label for query sample $i$ is a similarity-weighted average of the reference labels, with weights given by a softmax over similarity at scale $\tau$:
\begin{equation}
\label{eq:prediction}
\hat{\mathbf{y}}_i
  = \frac{\sum_{j \in R} \exp\!\big(\tau \cdot \mathrm{sim}(\mathbf{z}_i,\mathbf{z}_j)\big)\, \mathbf{y}_j}
         {\sum_{j \in R} \exp\!\big(\tau \cdot \mathrm{sim}(\mathbf{z}_i,\mathbf{z}_j)\big)},
\end{equation}
where $\mathbf{y}_j$ is the known label of reference sample $j$---at this stage, its one-hot CIFAR-100 label---so that $\hat{\mathbf{y}}_i$ is a distribution over the $C$ classes. Higher values of $\tau$ narrow the prediction toward the nearest references, while lower values draw on the reference set more evenly.

\textit{3. Loss calculation and backpropagation: }
Let $c_i$ denote the true class index of query sample $i$. Whereas Equation~\eqref{eq:prediction} uses only the \emph{reference} labels, the loss compares $\hat{\mathbf{y}}_i$ against the \emph{query} sample's own label, giving the cross-entropy
\begin{equation}
\label{eq:loss}
\mathcal{L} = -\frac{1}{|Q|} \sum_{i \in Q} \log \hat{y}_i^{(c_i)}.
\end{equation}
$\mathcal{L}$ is back propagated through Equation~\eqref{eq:prediction} and Equation~\eqref{eq:phi}, updating the weights of $\phi$ and the scale $\tau$ jointly. Gradients flow through the features of both the reference and query samples since both are produced by $\phi$.

\paragraph{Construction of the Split}
Sampling of $R$ and $Q$ must be class-stratified rather than uniform. Under a uniform split, most query samples have no reference sample of their own class in the same step, and Equation~\eqref{eq:prediction} cannot assign weight to a class it has not seen; the achievable loss is then floored and training makes no progress. We therefore draw $k$ subsets per step, each containing $s$ samples from every one of the $C$ classes, and rotate which subset serves as $R$, while the remaining $k-1$ form $Q$, so that every sample serves in both roles across the $k$ inner iterations. Drawing a fresh stratified sample each step keeps the cost per step fixed while letting the reference/query pairing vary across training. The values of $k$, $s$, and the optimization settings are given in Section~\ref{sec:experimental_setup}.

Because $\tau$ is calibrated against $|R|$ during training and then reused unchanged at detection time, where the reference set is the much smaller verified set $G$, a large mismatch between $|R|$ and $|G|$ leaves $\tau$ tuned for a setting the detector never encounters.
We therefore treat $|R|$ as a hyperparameter and report results for a range of $|R|$ in
\ifExtendedVersion
Appendix~\ref{sec:reference_size}.
\else%
Appendix B in the extended online version~\cite{han2026sage}.
\fi

\begin{algorithm}[ht]
\caption{SAGE feature extractor training}
\label{alg:feature_extractor}
\begin{algorithmic}[1]
\Require Dataset $D$ with $C$ classes, extractor $\phi$
\State Initialize $\tau \gets 1$
\For{$\textit{epoch} \gets 1$ to $N$}
  \For{each step}
    \State Draw $k$ class-stratified subsets $S_1, \ldots, S_k$          from $D$, each with $s$ images per class
    \For{$r \gets 1$ to $k$}
      \State $R \gets S_r$; \quad $Q \gets \bigcup_{l \ne r} S_l$
      \State $\mathbf{z} \gets \phi(\mathbf{x})$ for all
             $\mathbf{x} \in R \cup Q$
      \State Compute $\mathrm{sim}(\mathbf{z}_i, \mathbf{z}_j)$
             for $i \in Q$, $j \in R$ \\ \Comment{Eq.~\eqref{eq:sim}}
      \State Compute $\hat{\mathbf{y}}_i$ for $i \in Q$
             \Comment{Eq.~\eqref{eq:prediction}}
      \State Compute $\mathcal{L}$ \Comment{Eq.~\eqref{eq:loss}}
      \State Update $\phi$ and $\tau$ by gradient descent on $\mathcal{L}$ \\
             \Comment{gradients flow through both $R$ and $Q$}
    \EndFor
  \EndFor
\EndFor
\end{algorithmic}
\end{algorithm}

\subsection{Applying SAGE as a Poison Detector}
\label{sec:defense_application}

At detection time we reuse Equations~\eqref{eq:sim} and~\eqref{eq:prediction} with $g_j$ in place of $\mathbf{y_j}$, with two differences from training. First, $\phi$ and $\tau$ are frozen: no training or fine-tuning is performed on the target dataset. Second, the reference set is the verified set, $R = G$, and reference labels are no longer one-hot class vectors $\mathbf{y}_j$ but scalar poison indicators $g_j \in \{0,1\}$, with $g_j = 1$ for $j \in G_{\text{poison}}$ and $g_j = 0$ for $j \in G_{\text{clean}}$. So, each reference image retains its own class label as well as its poison status.

Every remaining training example takes the query role. For a candidate sample $\mathbf{x}_i$, Equation~\eqref{eq:prediction} returns
\begin{equation}
\hat{g}_i
  = \frac{\sum_{j \in G} \exp\!\big(\tau \cdot \mathrm{sim}(\mathbf{z}_i,\mathbf{z}_j)\big)\, g_j}
         {\sum_{j \in G} \exp\!\big(\tau \cdot \mathrm{sim}(\mathbf{z}_i,\mathbf{z}_j)\big)}
  \in [0,1],
\end{equation}
the similarity-weighted fraction of verified poisons among the reference set---an estimated poison probability. No loss is computed at this stage, since the candidate's true status is what is being inferred, and no parameters are updated.
We remove $\mathbf{x}_i$ from the training set when $\hat{g}_i \ge 0.5$. Examples in $G_{\text{clean}}$ are retained regardless, having already been verified.

\section{Experimental Setup}
\label{sec:experimental_setup}

\subsection{Datasets}
\label{sec:datasets}

We evaluate SAGE on CIFAR-10 and Tiny ImageNet, two standard benchmarks in the data-poisoning literature. CIFAR-10 contains 50{,}000 training images across 10 classes. Tiny ImageNet contains 100{,}000 training images across 200 classes.
Neither is used to train the feature extractor, which is trained on CIFAR-100 (Section~\ref{sec:extractor_setup}).

Experiments run on a server with a 64-core 2.4\,GHz AMD EPYC CPU and an NVIDIA RTX A5000 GPU (24\,GB). Training the feature extractor takes roughly 30 minutes.
Once trained, applying SAGE to a 50{,}000-image training set takes 4 minutes. The extractor is trained once and reused across every attack, dataset, and verified-set configuration reported below, so the marginal cost of each additional experiment is the 4-minute detection pass.

\subsection{Feature Extractor Training}
\label{sec:extractor_setup}

The extractor of Equation~\eqref{eq:phi} uses a ResNet-18 backbone trained from scratch on CIFAR-100~\cite{krizhevsky2009learning}, with a two-layer head of hidden width $h=512$ and output dimension $d=256$. Optimization uses SGD (learning rate 0.1, momentum 0.9, weight decay $5\times10^{-4}$, cosine-annealed) for 200 epochs, with random crops and horizontal flips. The scale $\tau$ is initialized to 1 and optimized jointly with the network weights.

Reference/query splits are drawn as described in Section~\ref{sec:training_phi}: at each step we sample $k=4$ class-stratified subsets of $s=2$ images per class, giving subsets of 200 images and a step size of $k \times s \times C = 800$ images, with roughly 62 steps per epoch. The reference set at training time therefore has $|R| = 200$.

\subsection{Verified Set}
\label{sec:verified_set}

Our default verified set on CIFAR-10 contains $c=2$ clean examples from each of the $K=10$ classes and $q=2$ confirmed poisons drawn from the base class, giving $|G| = 22$.
Poisons and clean examples are drawn from the poisoned training set itself, so $G \subset D'$; no external data is assumed.
Section~\ref{sec:ablations} varies both the total size of $G$ and its per-class composition.

\subsection{Victim Model Training}
\label{sec:training_mode}

After filtering, we train a victim model on the cleaned dataset. On CIFAR-10 we use white-box transfer learning with ResNet-18 for Feature Collision and Bullseye Polytope (40 epochs, SGD, learning rate 0.1, batch size 128), and train ResNet-18 from scratch for Gradient Matching (200 epochs, same optimizer settings). On Tiny ImageNet we train VGG-16 from scratch (200 epochs, learning rate 0.1, batch size 64).

For the trigger-based attacks in Section~\ref{sec:backdoor} we train ResNet-18 on CIFAR-10 from scratch with a single fixed recipe: SGD, learning rate 0.1, momentum 0.9, weight decay $5\times10^{-4}$, cosine-annealed, batch size 128, 100 epochs, random crop and flip. Holding this recipe constant across attacks ensures that differences in defended ASR and accuracy reflect the attack and defense rather than the training procedure; each attack's own configuration (trigger design, poison rate, target-class restriction) still follows its published setting.

\subsection{Poisoning Attacks}
\label{sec:poisoning_setup}

\paragraph{Triggerless attacks}
We follow the benchmark of Schwarzschild et al.~\cite{schwarzschild2021just} for the three triggerless attacks, adopting its hyperparameters, optimizers, and evaluation protocol. Results are reported over 20 trials with distinct base--target pairs. We evaluate $\epsilon \in \{8, 16, 22\}$, where $8$ and $16$ are common triggerless poisoning budgets and $22$ is included as an additional stronger-perturbation setting. Each setting generates 500 poisons for CIFAR-10 and 250 for Tiny ImageNet.

\textbf{Feature Collision}~\cite{shafahi2018poison} optimizes poisons to collide with the target in feature space, so that a decision boundary accommodating the poisons captures the target. The attacker knows the pretrained feature extractor the victim will fine-tune.

\textbf{Bullseye Polytope}~\cite{aghakhani2021bullseye} instead surrounds the target with a convex hull of poisons in feature space, placing the target near its center, which relaxes the collision constraint and transfers more reliably. The attacker again knows the pretrained extractor.

\textbf{Gradient Matching}~\cite{geiping2020witches} aligns the gradient of the training loss on the poisons with that of the adversarial objective, and is effective against models trained from scratch. Here the attacker knows only the victim's architecture and training procedure, not the resulting weights.

\paragraph{Backdoor attacks}
For backdoor attacks, we use each author's released implementation and default hyperparameters. For backdoor attacks, ASR is measured over the triggered test set rather than a single target sample; we also report clean accuracy.
Both are clean-label: the trigger is applied only to images already belonging to the target class, and no label is altered.

\textbf{Narcissus}~\cite{zeng2023narcissus} optimizes a global trigger without any knowledge of the victim's training set. The attacker trains a surrogate model on a public out-of-distribution dataset, appends the target class to it, and then searches for a perturbation that drives inputs strongly toward that class.

\textbf{Wicked Oddities}~\cite{nguyen2025wicked} keeps the trigger fixed and selects which images of the target-class to poison, motivated by the observation that samples contribute unequally to attack success. As the selection strategy is trigger-agnostic, we experiment with three instantiations: Wicked-BadNets, Wicked-Blended, and Wicked-SIG.

\subsection{Baseline Defenses}
\label{sec:baseline_defenses}

\textbf{Deep $k$-NN}~\cite{peri2020deep} flags a point whose label disagrees with the plurality label among its $k$ nearest neighbors in feature space. It is the most directly comparable baseline for triggerless attacks. It requires no verified set, only the class labels of the training data, which under a triggerless attack are correct by construction. Following the default setting, we set $k$ to the number of examples per class. 

\textbf{EPIC}~\cite{yang2022not} clusters examples in gradient space and drops points isolated from the rest during training, requiring no ground-truth information.

\textbf{Meta-Sift}~\cite{zeng2023meta} selects a clean subset of a user-specified size from a poisoned dataset. Following its default 1{,}000 sample selection budget, we use the complementary ranking as a filter and discard the 1{,}000 lowest-ranked images (leaving 49{,}000 for CIFAR-10 and 99{,}000 for Tiny ImageNet).

\textbf{Confusion Training}~\cite{qi2023towards} decouples benign correlations by training on poisoned data alongside randomly mislabeled clean samples, so that the model fits poisoned samples while failing to fit clean ones, exposing the poisons. 

\textbf{Activation Clustering}~\cite{chen2018detecting} clusters the final-layer activations of each class into two groups and flags the smaller one as poisoned, on the assumption that poisoned samples form a distinct cluster within their label.

\textbf{SPECTRE}~\cite{hayase2021spectre} estimates robust mean and covariance of each class's representations and whitens them, amplifying the directions in which poisoned samples deviate from the clean distribution. The samples are then ranked by a QUE score along these directions, and the highest-scoring ones are removed.

\textbf{FLARE}~\cite{hou2025flare} aggregates abnormal activations across all hidden layers rather than the final one, arguing that poison/benign separability varies by depth. It selects an adaptive subspace, splits the dataset into two clusters, and flags the more stable cluster as poisoned.

\subsection{Metrics}
\label{sec:metrics}

We report attack success rate (ASR) and clean test accuracy. For the three targeted attacks, ASR is the fraction of trials in which the single target image $x_t$ is classified as the attacker's chosen base label. Test accuracy is measured on the standard test split after training on the cleaned dataset.
We additionally report the fraction of the training set removed and the false-positive rate among removed examples (i.e., removed samples being benign), since a defense that removes a large share of clean data can suppress ASR while degrading the model it was meant to protect.

\section{Numerical Results}
\label{sec:results}

\subsection{Triggerless Attacks}
\label{sec:asr_results}

Table~\ref{tab:result} reports the post-defense attack success rate (ASR) and clean test accuracy on CIFAR-10 and Tiny ImageNet. Without a defense, ASR varies substantially across attacks. Bullseye Polytope succeeds in 95--100\% of trials, and Gradient Matching succeeds in 75--100\%. In contrast, Feature Collision succeeds in only 15--35\% of trials. Results for Feature Collision should therefore be interpreted relative to its lower undefended~ASR.

On CIFAR-10, SAGE achieves the lowest ASR in every setting, including ties, reducing ASR to 0\% for Feature Collision and Bullseye Polytope at all perturbation budgets. For Gradient Matching, it limits ASR to 0--5\%. EPIC also achieves 0\% ASR for Feature Collision at all three budgets. For Bullseye Polytope at $\epsilon=16$, SAGE achieves 0\% ASR, tied with Deep $k$-NN, compared with 15\% for EPIC, 35\% for Meta-Sift, and 70\% for Confusion Training.

The accuracy results are more mixed. Under transfer learning, EPIC and Deep $k$-NN achieve higher clean accuracy than SAGE, with ranges of 0.9455--0.9466 and 0.9389--0.9481, respectively, compared with 0.9258--0.9428 for SAGE. 
The ordering reverses under from-scratch training for Gradient Matching, where SAGE retains 0.8980--0.9033 against 0.8787--0.8806 for EPIC, though Confusion Training is best at 0.9040--0.9122. Relative to the undefended models, SAGE costs 0.3--5.1 percentage points, with the largest reductions under from-scratch training.

\paragraph{Tiny ImageNet}

We evaluate Bullseye Polytope on Tiny ImageNet using VGG-16. SAGE reduces ASR from 30\% to 10\% and slightly improves accuracy from 0.6153 to 0.6169. It is the only defense that avoids an accuracy loss in this setting. Deep $k$-NN reduces ASR to 20\%, but its accuracy falls to 0.5584. EPIC leaves ASR unchanged at 30\% and reduces accuracy to 0.5569.

\begin{table*}[!ht]
\centering
\scriptsize
\caption{Post-defense attack success rate (ASR) and clean test accuracy (Acc.) for clean-label triggerless poisoning attacks on CIFAR-10 and Tiny ImageNet. Results are averaged over 20 poisoned datasets for each attack and perturbation budget $\epsilon$. ASR is the fraction of successful attacks across the 20 poisoned datasets, while Acc. is reported as mean $\pm$ standard deviation. For each attack, the best result among the defenses is shown in bold, and the second-best result is underlined.}
\label{tab:result}
\resizebox{\textwidth}{!}{%
\begin{tabular}{ll|cc||cc|cc|cc|cc|cc}
\toprule
\multicolumn{2}{c|}{\textbf{Attack}} & 
\multicolumn{2}{c||}{\textbf{No Defense}}
& \multicolumn{2}{c|}{\textbf{SAGE (Ours)}}
& \multicolumn{2}{c|}{\textbf{Deep $k$-NN}}
& \multicolumn{2}{c|}{\textbf{Meta-Sift}}
& \multicolumn{2}{c|}{\textbf{EPIC}}
& \multicolumn{2}{c}{\textbf{Confusion Training}}\\
{Size} & {Method} & Acc. ($\uparrow$) & ASR ($\downarrow$) & Acc. ($\uparrow$) & ASR ($\downarrow$) & Acc. ($\uparrow$) & ASR ($\downarrow$) & Acc. ($\uparrow$) & ASR ($\downarrow$)  & Acc. ($\uparrow$) & ASR ($\downarrow$) & Acc. ($\uparrow$) & ASR ($\downarrow$)  \\ 
\midrule
\multicolumn{14}{c}{\textbf{CIFAR-10}} \\
\midrule
{$\epsilon$ = 8}& FC & 0.9460$\pm$0.0008 & 15\% & 0.9258$\pm$0.0017 & \textbf{0\%} & 0.9389$\pm$0.0034 & \textbf{0\%} & 0.9233$\pm$0.0003 & 15\% & \underline{0.9465$\pm$0.0008} & \textbf{0\%} & \textbf{0.9474$\pm$0.0042} & 5\% \\
& BP  & 0.9457$\pm$0.0007 & 100\%  & 0.9366$\pm$0.0012 & \textbf{0\%} & \textbf{0.9471$\pm$0.0002} & 20\%  &0.9155$\pm$0.0002 & 25\% & \underline{0.9466$\pm$0.0008} & \underline{10\%} & 0.9416$\pm$0.0046  & 45\%\\
& GM  & 0.9487$\pm$0.0006 & 75\% & \underline{0.8980$\pm$0.0025} & \textbf{0\%} &0.8973$\pm$0.0123 & \underline{10\%} & 0.8907$\pm$0.0009 & 20\% & 0.8787$\pm$0.0048 & \underline{10\%} & \textbf{0.9040$\pm$0.0065} & \underline{10\%}\\ \midrule
{$\epsilon$ = 16} & FC & 0.9456$\pm$0.0008 & 25\% & 0.9392$\pm$0.0024 & \textbf{0\%} & 0.9426$\pm$0.0032 & 10\% & 0.9255$\pm$0.0003 & 25\% & \textbf{0.9465$\pm$0.0006} & \textbf{0\%} & \underline{0.9451$\pm$0.0032} & 5\%\\
& BP& 0.9458$\pm$0.0009 & 100\% & 0.9375$\pm$0.0031  & \textbf{0\%} & \underline{0.9389$\pm$0.0015} & \textbf{0\%} & 0.9242$\pm$0.0002& 35\% & \textbf{0.9455$\pm$0.0008} & 15\% & 0.9334$\pm$0.0033 & 70\%\\
& GM & 0.9486$\pm$0.0008 & 95\% & 0.8986$\pm$0.0021 & \textbf{5\%} & \underline{0.9007$\pm$0.0032} & 10\% & 0.8952$\pm$0.0009  & 40\%& 0.8794$\pm$0.0035 & \textbf{5\%} & \textbf{0.9051$\pm$0.0047} & 15\%\\
\midrule
{$\epsilon$ = 22} & FC & 0.9455$\pm$0.0008  & 35\% & 0.9428$\pm$0.0048 & \textbf{0\%} & \textbf{0.9481$\pm$0.0020} & 10\% & 0.9258$\pm$0.0003 & 35\% & \underline{0.9465$\pm$0.0009} & \textbf{0\%} & 0.9417$\pm$0.0034 & 20\%\\
& BP  & 0.9458$\pm$0.0007 & 95\% & 0.9428$\pm$0.0041 & \textbf{0\%} & \underline{0.9477$\pm$0.0032} & 10\% & 0.9258$\pm$0.0002 & 40\% & 0.9465$\pm$0.0008 & \textbf{0\%} & \textbf{0.9500$\pm$0.0029} & 25\%\\
& GM  & 0.9491$\pm$0.0009 & 100\% & \underline{0.9033$\pm$0.0029} & \textbf{5\%} & 0.8985$\pm$0.0039 & \textbf{5\%} & 0.8837$\pm$0.0011 & 60\% & 0.8806$\pm$0.0023 & \textbf{5\%} & \textbf{0.9122$\pm$0.0064} & 15\%\\
\midrule
\multicolumn{14}{c}{\textbf{Tiny ImageNet}} \\ \midrule
{$\epsilon$ = 8} & BP & 0.6153$\pm$0.0047 & 30\% & \textbf{0.6169$\pm$0.006} & \textbf{10\%} & 0.5584$\pm$0.0051 & \underline{20\%} & 0.5869$\pm$0.0021 & 50\% & 0.5569$\pm$0.0047 & 30\% & \underline{0.6025$\pm$0.0057} & 25\%\\
\bottomrule
\end{tabular}}
\end{table*}

\subsection{Backdoor Attacks}
\label{sec:backdoor}
Table~\ref{tab:results2} reports results for the four clean-label backdoor attacks on CIFAR-10. Each poisons 10\% of an attacker-selected target class, corresponding to 1\% of the full training set. We measure ASR over the 9{,}000 test images outside the target class after applying the trigger.

SAGE reduces ASR for all four attacks, reaching 0.1310 on Narcissus and 0.2720 on Wicked-BadNets. Wicked-Blended has the highest residual ASR under SAGE at 0.4620, followed by Wicked-SIG at 0.3400. Clean accuracy decreases by 2.40 percentage points on Narcissus and by at most 0.50 percentage points on the Wicked variants. On Narcissus, Activation Clustering~\cite{chen2018detecting}, SPECTRE~\cite{hayase2021spectre}, and Confusion Training achieve lower ASR than SAGE, but also lower clean accuracy. On all three Wicked variants, SAGE achieves lower ASR than these defenses while maintaining clean accuracy above 0.94.

FLARE~\cite{hou2025flare} achieves the lowest ASR on the three Wicked variants, but its clean accuracy ranges from 0.4586 to 0.5507 across the four attacks, compared with 0.9281--0.9494 for SAGE. These results highlight the trade-off between attack suppression and clean accuracy: although SAGE does not achieve the lowest ASR overall, it substantially reduces ASR across all four attacks while retaining higher clean accuracy than FLARE.

\begin{table*}[ht]
    \centering
    \caption{Post-defense clean test accuracy (Acc.) and attack success rate (ASR) for four clean-label backdoor attacks on CIFAR-10. Results are mean $\pm$ std over 10 runs. For each attack, the best result among the defenses is shown in bold, and the second-best result is underlined. Note that FLARE's low ASR comes at the cost of excessively low clean accuracy.}
    \label{tab:results2}
    \scriptsize
    \resizebox{\textwidth}{!}{%
    \begin{tabular}{l|cc||cc|cc|cc|cc|cc}
        \toprule
        \multirow{2.5}{*}{\textbf{Attack}} 
            & \multicolumn{2}{c||}{\textbf{No Defense}}
            & \multicolumn{2}{c|}{\textbf{SAGE (Ours)}}
            & \multicolumn{2}{c|}{\textbf{Activation Clustering}}
            & \multicolumn{2}{c|}{\textbf{SPECTRE}}
            & \multicolumn{2}{c|}{\textbf{Confusion Training}}
            & \multicolumn{2}{c}{\textbf{FLARE}}
            \\
            & Acc. ($\uparrow$) & ASR ($\downarrow$) & Acc. ($\uparrow$) & ASR ($\downarrow$) & Acc. ($\uparrow$) & ASR ($\downarrow$) & Acc. ($\uparrow$) & ASR ($\downarrow$) & Acc. ($\uparrow$) & ASR ($\downarrow$) & Acc. ($\uparrow$) & ASR ($\downarrow$) \\
        \midrule
\multicolumn{13}{c}{CIFAR-10} \\
        \midrule
        Narcissus
            & 0.9521$\pm$0.0004  & 0.9120$\pm$0.0010
            & \textbf{0.9281$\pm$0.0030} & 0.1310$\pm$0.0017
            & 0.9023$\pm$0.0020 & \textbf{0.0431$\pm$0.0025}
            & 0.9045$\pm$0.0006 & 0.1235$\pm$0.0290
            & \underline{0.9149$\pm$0.0010} & 0.0987$\pm$0.0061
            & 0.5364$\pm$0.0294 & \underline{0.0452$\pm$0.0011}
            \\
        Wicked-BadNets
            & 0.9467$\pm$0.0006 & 0.8970$\pm$0.0035
            & \textbf{0.9494$\pm$0.0022} & \underline{0.2720$\pm$0.0048}
            & 0.9469$\pm$0.0006 & 0.8915$\pm$0.0067
            & 0.9425$\pm$0.0055 & 0.8959$\pm$0.0174
            & \underline{0.9475$\pm$0.0049} & 0.8065$\pm$0.0074
            & 0.5507$\pm$0.0392& \textbf{0.0699$\pm$0.0469} 
            \\
        Wicked-Blended
            & 0.9471$\pm$0.0002 & 0.9383$\pm$ 0.0007
            & \textbf{0.9467$\pm$0.0007} & \underline{0.4620$\pm$0.0030}
            & \underline{0.9447$\pm$0.0029} & 0.9381$\pm$0.0014
            & 0.9428$\pm$0.0047 & 0.5608$\pm$0.0758
            & 0.9408$\pm$0.0042 & 0.5675$\pm$0.0054
            & 0.4586$\pm$0.0040 & \textbf{0.1023$\pm$0.0762}
            \\
        Wicked-SIG
            & 0.9451$\pm$0.0007 & 0.9372$\pm$0.0018
            & 0.9401$\pm$0.0022 & \underline{0.3400$\pm$0.0157}
            & \underline{0.9407$\pm$0.0056} & 0.9382$\pm$0.0262
            & 0.9400$\pm$0.0054 & 0.7613$\pm$0.0475
            & \textbf{0.9452$\pm$0.0041} & 0.7612$\pm$0.0038
            & 0.4941$\pm$0.0354 & \textbf{0.0717$\pm$0.0206}
            \\
        \bottomrule
    \end{tabular}
}
\end{table*}

\subsection{Precision of Removal}
\label{sec:removal_results}

Table~\ref{tab:number_of_poison} reports the total number of removed samples and the number of true poisons among them. These results reveal substantial differences in removal precision.

On CIFAR-10, SAGE recovers more than 480 of the 500 poisons while removing 769.2--995.1 images in total, or 1.5--2.0\% of the training set, with 35.4--50.8\% false positives. The baselines trade off differently. Meta-Sift removes 1{,}000 images in every setting but recovers at most 105.1 poisons, a recall up to 21\%. EPIC removes 9{,}189.5--9{,}290 images under transfer learning and about 19{,}405.1--20{,}174.2, around 38.8--40.3\% of the dataset, from scratch, in the latter case recovering fewer than 100 poisons. Deep $k$-NN removes the fewest images, 205.3--406.4, but 46.1--90.7\% of those are false positives.

For Gradient Matching at $\epsilon=16$, SAGE and EPIC both achieve an ASR of 5\%, but remove an average of 896.9 and 20{,}174.2 images, respectively. Thus, the same ASR can conceal a large difference in the amount of training data discarded, motivating the reporting of removal counts alongside ASR.

The pattern differs on Tiny ImageNet. SAGE removes 187 images, including 126 of the 250 poisons. This result corresponds to 50.4\% recall and a 32.6\% of removed examples being benign, the best removal precision among the evaluated defenses. Despite removing only about half of the poisons, SAGE reduces ASR from 30\% to 10\%. This result suggests that the attack depends on retaining a sufficiently large fraction of its poisons.

\begin{table}[t]
\centering
\scriptsize
\setlength{\tabcolsep}{3pt}
\caption{Examples identified as poisoned by each defense. TP denotes the number of true poisons identified, and Total denotes the total number of examples removed. The datasets contain 500 poisons on CIFAR-10 and 250 poisons on Tiny ImageNet. Results are mean $\pm$ std over 20 runs.}
\label{tab:number_of_poison}
\begin{adjustbox}{max width=\columnwidth}
\begin{tabular}{l c| cc|cc|cc|cc}
\toprule
\multicolumn{2}{c|}{\textbf{Attack}} & \multicolumn{2}{c|}{\textbf{SAGE}} & \multicolumn{2}{c|}{\textbf{Deep $k$-NN}} & \multicolumn{2}{c|}{\textbf{Meta-Sift}} & \multicolumn{2}{c}{\textbf{EPIC}} \\
Size & Method & TP ($\uparrow$) & Total & TP ($\uparrow$) & Total & TP ($\uparrow$) & Total & TP ($\uparrow$) & Total \\
\midrule
\multicolumn{10}{c}{\textbf{CIFAR-10}} \\ \midrule
$\epsilon = 8$  & FC & \ms{489.6}{23.4} & \ms{995.1}{34.1} & \ms{34.0}{33.4}   & \ms{220.8}{31.0}  & \ms{37.1}{2.4}  & \ms{1{,}000.0}{0.0} & \ms{388.1}{54.1} & \ms{9{,}191.8}{220.9} \\
& BP & \ms{500.0}{0.0}  & \ms{847.2}{14.5} & \ms{64.5}{53.0}   & \ms{251.8}{51.2}  & \ms{69.5}{18.4}  & \ms{1{,}000.0}{0.0} & \ms{448.2}{50.4} & \ms{9{,}242.5}{148.1} \\
& GM & \ms{482.1}{14.3} & \ms{844.3}{13.7} & \ms{19.0}{14.2}   & \ms{205.3}{13.4}  & \ms{24.3}{4.0}  & \ms{1{,}000.0}{0.0} & \ms{58.2}{30.2}  & \ms{19{,}405.1}{446.5} \\
\midrule
$\epsilon = 16$ & FC & \ms{496.9}{2.5}  & \ms{769.2}{13.8} & \ms{85.4}{69.2}   & \ms{273.1}{68.0}  & \ms{42.3}{3.4}  & \ms{1{,}000.0}{0.0} & \ms{454.0}{35.7} & \ms{9{,}189.5}{205.7} \\
& BP & \ms{495.5}{2.2}  & \ms{831.6}{14.9} & \ms{175.6}{113.3} & \ms{363.2}{111.5} & \ms{84.4}{49.6}  & \ms{1{,}000.0}{0.0} & \ms{480.7}{69.0} & \ms{9{,}237.0}{188.7} \\
& GM & \ms{480.4}{13.5} & \ms{896.9}{23.7} & \ms{39.0}{34.8}   & \ms{225.3}{35.2}  & \ms{50.6}{28.4}  & \ms{1{,}000.0}{0.0} & \ms{59.5}{35.2}  & \ms{20{,}174.2}{648.9} \\
\midrule
$\epsilon = 22$ & FC & \ms{496.2}{1.1}  & \ms{830.7}{14.4} & \ms{115.2}{92.5}  & \ms{303.2}{91.1}  & \ms{29.7}{5.9}  & \ms{1{,}000.0}{0.0} & \ms{374.2}{28.9} & \ms{9{,}196.3}{215.4} \\
& BP & \ms{499.8}{0.5}  & \ms{858.2}{8.9}  & \ms{219.2}{123.9} & \ms{406.4}{122.1} & \ms{105.1}{59.8} & \ms{1{,}000.0}{0.0} & \ms{426.6}{75.0} & \ms{9{,}290.0}{177.8} \\
& GM & \ms{496.2}{1.5}  & \ms{827.1}{13.8} & \ms{56.4}{30.5}   & \ms{242.4}{29.9}  & \ms{59.2}{35.9}  & \ms{1{,}000.0}{0.0} & \ms{57.4}{30.4}  & \ms{20{,}002.4}{945.1} \\
\midrule
\multicolumn{10}{c}{\textbf{Tiny ImageNet}} \\ \midrule
$\epsilon = 8$  & BP & \ms{126.0}{38.5} & \ms{187.0}{50.4} & \ms{77.7}{11.8}   & \ms{220.4}{18.4}   & \ms{3.3}{3.4}   & \ms{1{,}000.0}{0.0} & \ms{87.4}{5.3}  & \ms{40{,}036.0}{945.1} \\
\bottomrule
\end{tabular}
\end{adjustbox}
\end{table}

\subsection{Composition of the Verified Set}
\label{sec:ablations}

We next vary the composition of the verified set while holding all other components fixed. The feature extractor is trained once and reused across configurations. Each configuration therefore requires only an additional detection pass.

\begin{figure}
    \centering
    \includegraphics[width=\linewidth]{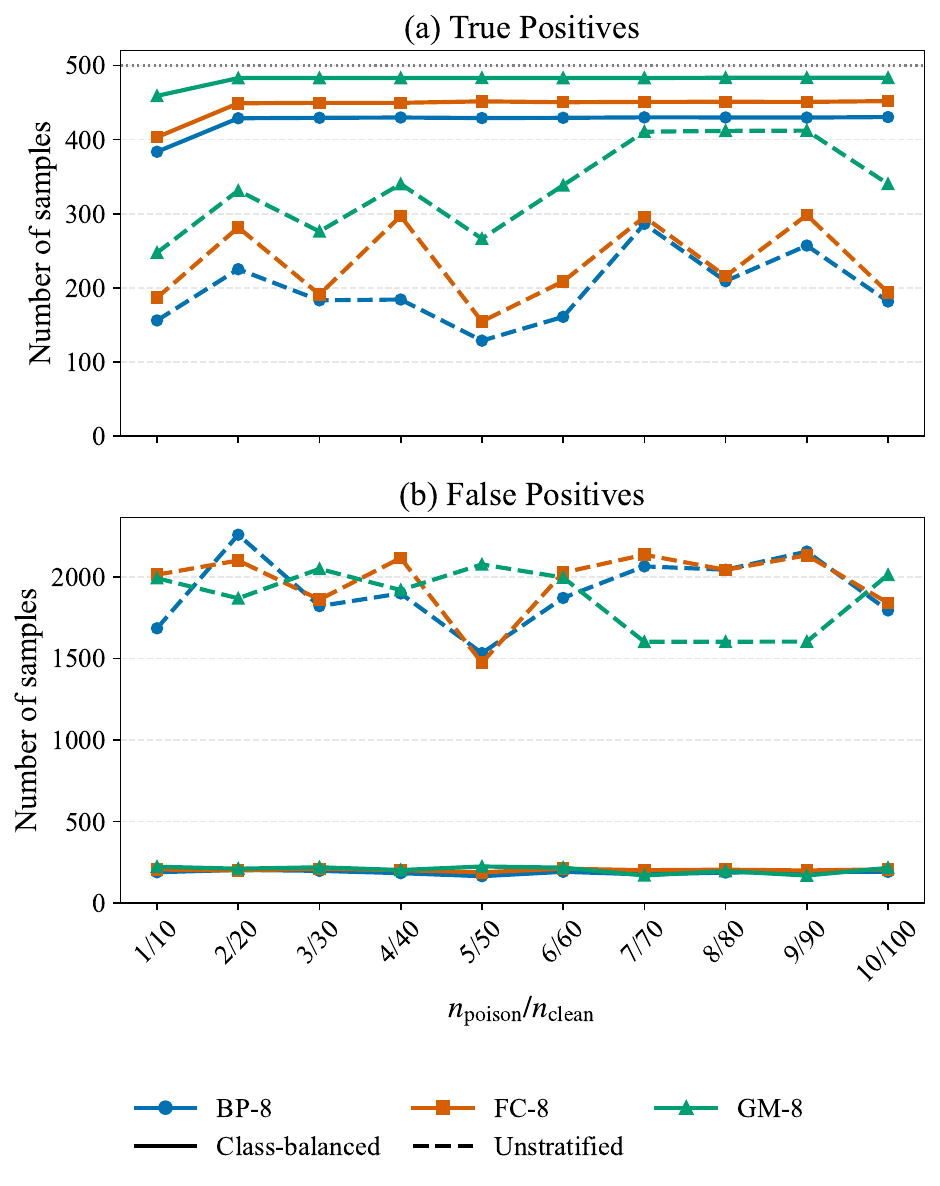}
    \vspace{-2em}
    \caption{Effect of verified-set composition on poison recall and false positives. The upper panel reports the number of poisoned images identified, and the lower panel reports the number of clean images removed. Solid lines denote class-balanced sampling of verified clean examples. Dashed lines denote unstratified random sampling.}
    \label{fig:balanced_vs_unstratified}
\end{figure}

Figure~\ref{fig:balanced_vs_unstratified} compares two verified-set sampling strategies as the number of verified poisons increases. With class-balanced clean examples, SAGE recovers most poisons using only two verified poisons for BP-8, FC-8, and GM-8. With unstratified clean examples, approximately five verified poisons are needed to achieve comparable recall.

The difference is more pronounced for false positives. Class-balanced sampling removes approximately 180--220 clean images. Unstratified sampling removes approximately 1{,}500--2{,}200 clean images, even when it achieves lower recall. These results show that balanced coverage of the clean classes is important for controlling false positives.

\section{Discussion and Conclusion}
\label{sec:conclusion}

\subsection{Adaptive Attacks}

We consider how an adaptive attacker might target SAGE directly. An adaptive attacker could attempt to lower the similarity $\mathrm{sim}(\mathbf{z}_i, \mathbf{z}_j)$ between each poison and the verified poisons in $G_{\text{poison}}$. Alternatively, the attacker could increase its similarity to the examples in $G_{\text{clean}}$. Either strategy could reduce the poison probability returned by Equation~\eqref{eq:prediction}. However, two aspects of our setting complicate such attacks.

First, $\phi$ is trained on CIFAR-100, disjoint from the attacked datasets. It is also not exposed to the poisoning. An attacker without access to the weights of $\phi$ cannot directly compute or differentiate through the similarity function used by SAGE.

Second, even a white-box attacker with access to $\phi$ does not know which examples will form $G$. The verified set is built independently of the attack, typically by inspecting a small sample of training data after poisoning has occurred. The attacker can therefore target only the distribution of clean and poisoned embeddings, but not the specific reference examples used for detection.
This separation is part of our threat model, not a general robustness guarantee; we make no claim against
an attacker who knows $G$ when constructing the poisons.

A more fundamental line of attack targets the assumption that verified poisons are representative of unverified ones. SAGE relies on poisoned examples occupying related regions of the feature space. An attacker could challenge this by increasing diversity among the poisons, sacrificing the effectiveness of each individual poison in exchange for lower similarity across poisons. Unlike the strategies above, this one is not blocked by the attacker's uncertainty about $\phi$ or $G$, since it targets the central premise of SAGE rather than the specific reference examples used at detection time.

Evaluating these adaptive strategies is an important direction for future work. One approach is to add an evasion term to the Bullseye Polytope or Gradient Matching objective, penalizing proximity to a verified set. Another is to construct diversity-regularized variants of these attacks and measure the resulting trade-off between attack effectiveness and detectability.

\subsection{Limitations}
\label{sec:limitations}

\paragraph{Assumption of correct verified set}
SAGE has a weaker form of the assumption made by the base-set defenses: we do not require a large verified set to be pure, but we do assume that examples identified as clean are indeed clean. A poison incorrectly included in $G_{\text{clean}}$ would provide an erroneous reference for the similarity-weighted prediction. The assumption is easier to satisfy at our scale, since certifying a few dozen examples is more tractable than certifying the hundreds or thousands used elsewhere, but our experiments do not evaluate contamination of $G_{\text{clean}}$ at any scale. Measuring how performance degrades under such contamination is an important robustness experiment left for future work.

\paragraph{Cross-dataset reuse of the feature extractor}
We train $\phi$ on CIFAR-100 and reuse it without fine-tuning on CIFAR-10 and Tiny ImageNet. This design avoids retraining for each dataset, but the learned embedding reflects the class structure of CIFAR-100 rather than that of the target dataset. Our experiments show that the embedding separates clean and poisoned examples in the evaluated settings, though the method does not guarantee separation elsewhere. Performance may decline when the target domain differs from natural images, such as in medical imaging, and extending the method to non-visual modalities would require separate evaluation.

\subsection{Conclusion}

This work proposes SAGE, a defense against clean-label poisoning that assumes a small set of examples verified as clean or poisoned, rather than a large base set presumed clean, and classifies the remaining data by similarity to them in a learned feature space. On CIFAR-10, SAGE holds attack success to at most 5\% against Feature Collision, Bullseye Polytope, and Gradient Matching while recovering at least 480 of the 500 poisons, on average. On Tiny ImageNet, SAGE reduces Bullseye Polytope ASR from 30\% to 10\% while recovering 126 of 250 poisons. It also suppresses four clean-label backdoor attacks without the accuracy loss incurred by the one baseline reaching comparable success rates. What separates it from existing defenses is how little it discards: against Gradient Matching on CIFAR-10, SAGE removes fewer than 900 images on average at every tested perturbation budget, whereas EPIC removes more than 19{,}000. Knowing even a handful of poisoned examples is thus disproportionately valuable, and verification effort is better spent identifying examples of both kinds than assembling a larger set assumed to be clean.

\section*{Acknowledgments}

This work was partially supported by the Army Research Office (W911NF-25-1-0059) and Office of Naval Research (N000142412663).

\bibliographystyle{IEEEtran}
\bibliography{main}

\ifExtendedVersion
\clearpage
\appendix
\section{Feature-Extractor Hyperparameters}
\label{app:hyperparameters}

The feature extractor $\phi$ is trained once on CIFAR-100 and reused for all detection experiments, so its two design choices---the output dimension $d$ and the training reference-set size $|R|$---are fixed before any poisoned data is seen. We select both by measuring cross-entropy loss and accuracy on the clean CIFAR-100 test set under the similarity-weighted prediction rule in Equation~\eqref{eq:prediction}, the same rule used at detection time.

\subsection{Feature Extractor Output Dimension}
\label{sec:hyperparameter_search}

Table~\ref{tab:featureextractor} evaluates four output dimensions for the feature extractor. We report cross-entropy loss and accuracy on the clean CIFAR-100 test set using the reference-query prediction rule in Equation~\eqref{eq:prediction}. Performance is not monotonic in the output dimension. The dimension $d=256$ achieves the lowest loss of 2.0417 and the highest accuracy of 0.7328. Increasing the dimension to 512 slightly degrades both metrics. This result suggests that the projection provides a useful bottleneck rather than simply limiting model capacity. We therefore use $d=256$ in all experiments. The learned scale $\tau$ ranges from 13.03 to 13.56 across the evaluated dimensions and is largest at $d=256$.

\begin{table}[ht]
\centering
\caption{Effect of the feature-extractor output dimension $d$. Cross-entropy loss and accuracy are measured on the clean CIFAR-100 test set using the similarity-weighted prediction rule in Equation~\eqref{eq:prediction}.}
\label{tab:featureextractor}
\begin{tabular}{c c c c}
\toprule
\textbf{Dimension} & \textbf{Scale} $\tau$ & \textbf{Loss} & \textbf{Accuracy} \\ 
\midrule
64  & 13.5630 & 2.3252 & 0.7020 \\ 
128 & 13.0431 & 2.2413 & 0.7146 \\ 
\textbf{256} & 13.0456 & \textbf{2.0417} & \textbf{0.7328} \\ 
512 & 13.0370 & 2.1955 & 0.7242 \\ 
\bottomrule
\end{tabular}
\end{table}

\subsection{Reference Set Size}
\label{sec:reference_size}

The scale $\tau$ is learned during training and reused during detection. The size of the training reference set $R$ is therefore an important design choice. Table~\ref{tab:reference_size} shows that accuracy is highest at $|R|=200$ and decreases for both smaller and larger reference sets.

The result for $|R|=22$ is particularly informative. With 100 classes, such a small reference set covers only about one fifth of the classes. Most queries therefore lack a same-class reference. In response, the learned value of $\tau$ approaches zero, and the prediction rule becomes close to a uniform average. These results indicate that class coverage, rather than reference-set size alone, is important. At $|R|=200$, the extractor receives an average of two references per class. We therefore use $|R|=200$.

\begin{table}[ht]
\centering
\caption{Effect of the reference-set size $|R|$ used during feature-extractor training. Accuracy is measured on the clean CIFAR-100 test set using the similarity-weighted prediction rule in Equation~\eqref{eq:prediction}.}
\label{tab:reference_size}
\begin{tabular}{c c c}
\toprule
$|R|$ & \textbf{Scale} $\tau$ & \textbf{Accuracy} \\
\midrule
22  & 0.0000 & 0.5166 \\
100 & 1.2895 & 0.6386 \\
\textbf{200} & 3.5260 & \textbf{0.7350} \\
500 & 2.0171 & 0.6867 \\
\bottomrule
\end{tabular}
\end{table}

\fi

\end{document}